\documentclass{article}
\usepackage{graphicx}
\usepackage{amsfonts}
\usepackage{amsmath}
\usepackage[font={small,it}]{caption}
\usepackage{caption}
\usepackage{multicol}
\usepackage{subcaption}
\DeclareMathOperator*{\KL}{KL}
\usepackage[final]{corl_2017} 
\title{Improved Adversarial Systems for 3D Object Generation and Reconstruction}

\author{
  Edward J. Smith\\
  Department of Computer Science\\
  McGill University 
  Canada\\
  \texttt{edward.smith@mail.mcgill.ca} \\
  \And
  David Meger\\
  Department of Computer Science\\
  McGill University 
  Canada\\
  \texttt{dmeger@cim.mcgill.ca } \\
}

\begin{document}
\maketitle


\begin{abstract}
	This paper describes a new approach for training generative adversarial networks (GAN) to understand the detailed 3D shape of objects. While GANs have been used in this domain previously, they are notoriously hard to train, especially for the complex joint data distribution over 3D objects of many categories and orientations. Our method extends previous work by employing the Wasserstein distance normalized with gradient penalization as a training objective. This enables improved generation from the joint object shape distribution. Our system can also reconstruct 3D shape from 2D images and perform shape completion from occluded 2.5D range scans. We achieve notable quantitative improvements in comparison to existing baselines. 
\end{abstract}



\section{Introduction} Understanding the 3D shape of objects in the world is a crucial step for many areas of robotics. Across object categories, shape is used for classification and to comprehend affordances. Within categories, fine shape details and textures contribute to successful manipulation. The complexity of the combined shape distribution has limited the success of parametric models historically used for the task. Recently, techniques for unsupervised learning of deep networks trained on large data-sets of synthetic 3D shapes have produced models capable of representing the detailed shape space \cite{DBLP:journals/corr/RezendeEMBJH16, DBLP:journals/corr/SharmaGF16}. This allows the generation of strikingly realistic sample shapes for simulation or visualization as well as the interpretation of robot sensory data.

\begin{figure}[t]
\centering{}
\includegraphics[width=.6\textwidth]{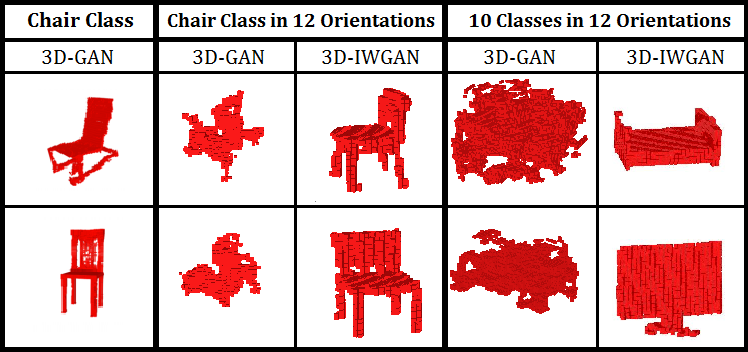}
\caption{First column: images of objects produced by the 3D-GAN system copied from \citep{3DGAN}. Last 4 columns: object produced by the 3D-GAN and 3D-IWGAN systems when trained on the ModelNet10 dataset.}
\label{comparison}
\end{figure}

This paper improves and extends a recent method known as 3DGAN \citep{3DGAN}, which has shown the ability to generate realistic samples of 3D object shapes, to categorize 3D shapes, and to reconstruct 3D shape from 2D images. 3DGAN is based upon the original generative adversarial network (GAN)~\citep{GAN_Original} architecture and training approach, which is well-known to suffer from instability. While 3DGAN is proficient at producing high quality objects from single classes, we have observed that it is difficult to train on distributions involving multiple distinct object classes in varied poses, as can be seen in Figure \ref{comparison}, leading the authors to primarily report results from independent models, each trained only on a single object category. We have been motivated to pursue joint training over a mixture of categories, following the goal of completely data-driven modeling. Can this complex shape distribution be captured without category label supervision?

The GAN architecture assigns each object it has experienced to a point in a latent space. The remainder of this space can allow generation of novel realistic objects, but only if the GAN training has converged successfully to the data distribution. For simple data distributions, such as objects from a single category, a simple network structure such as linear interpolation may suffice. Complex data distributions, such as those formed by the sharp transitions between diverse object categories, lead to harder training problems. 


In order to capture these wider and more complicated distributions, we propose the use of the recently proposed Wasserstein training objective with gradient penalty, a technique that has been successful in training GANs for other complex tasks. We call our resulting shape model the 3D Improved Wasserstein Generative Adversarial Network (3D-IWGAN). Our results show that this approach improves the stability of training for complex distributions. We are able to generate multiple object categories from multiple viewpoints using a single joint shape network. We show that 3D-IWGAN can be applied to reconstructing  3D shape from 2D images by integrating our training with a Variational Auto-Encoder (VAE) \citep{VAE}. This leads to our new 3D-VAE-IWGAN system which achieves state of the art accuracy for the task, achieving an mean average precision score of 61.7 percent on the IKEA dataset, an 8.6 percent increase. This extended system relies on novel update rules to synchronize between encoding, generation and discrimination tasks, which stabilized our model's learning and were key in establishing their convergence. Our 3D-VAE-IWGAN system also allows us to perform shape completion, where full 3D shape is determined based on a 2.5D range scan, effectively removing the occlusion due to the sensor's geometry. In order to ensure reproducible experimental comparison, our code for all of these systems is publicly available on a GitHub repository\footnote{https://github.com/EdwardSmith1884/3D-IWGAN}. 


\section{Related Works}

The pioneering work of Roberts \cite{roberts1963} and Marr \cite{Marr:1982:VCI:1095712} exemplify early shape models composed of 3D geometric primitives with recognizable 2D properties. Similarly intuitive properties are still in common use, such as symmetric parts \cite{KimShapeCompletionSIGRAPH2015}, skeletons \cite{Sundar:2003:SBS:829510.830339} and CAD wire-frames \cite{deepCNNLiCVPR2017}.

Recently, generating distributions based upon the parameters of a complex function learned from data have been utilized to represent shapes without the need for any prior structural knowledge \cite{karAndMalikShapeReconstructionCVPR2015}. Recent examples include: GP-LVMs \cite{DameAndReidShapePriorsForReconstructionCVPR2013}, Convolutional Neural Networks \cite{broxChairCNN2015}, Recurrent Neural Networks \cite{3dr2n2}, Deep Belief Networks \cite{Wu3DShapenets}, Generative Adversarial Networks \cite{laplacianPyramidGAN,3DGAN} and Deep Convolutional Auto-encoders \cite{old}. The work that we extend upon in this paper, 3DGAN \cite{3DGAN}, falls in this category. Several of these shape models have been shown in practical applications such as interactive editing \cite{Modelling} and reconstructing room geometry \cite{3dgp,Sundar:2003:SBS:829510.830339,izadinia2017im2cad} and depth \cite{DepthInfo} from a single image can which act as import cues for understanding scenes at a higher level \cite{3DInfo,liIROS2017}. 

\section{Methods}

As shown in Figure 2, our 3D-IWGAN architecture utilizes voxelized objects from a data set to train deep generator and discriminator networks in tandem, with the goal of generating realistic 3D object shapes. Our discussion will focus on the improvements we made to the GAN training process that were crucial in capturing this complex distribution.

\begin{figure}[ht]
\includegraphics[width=\textwidth]{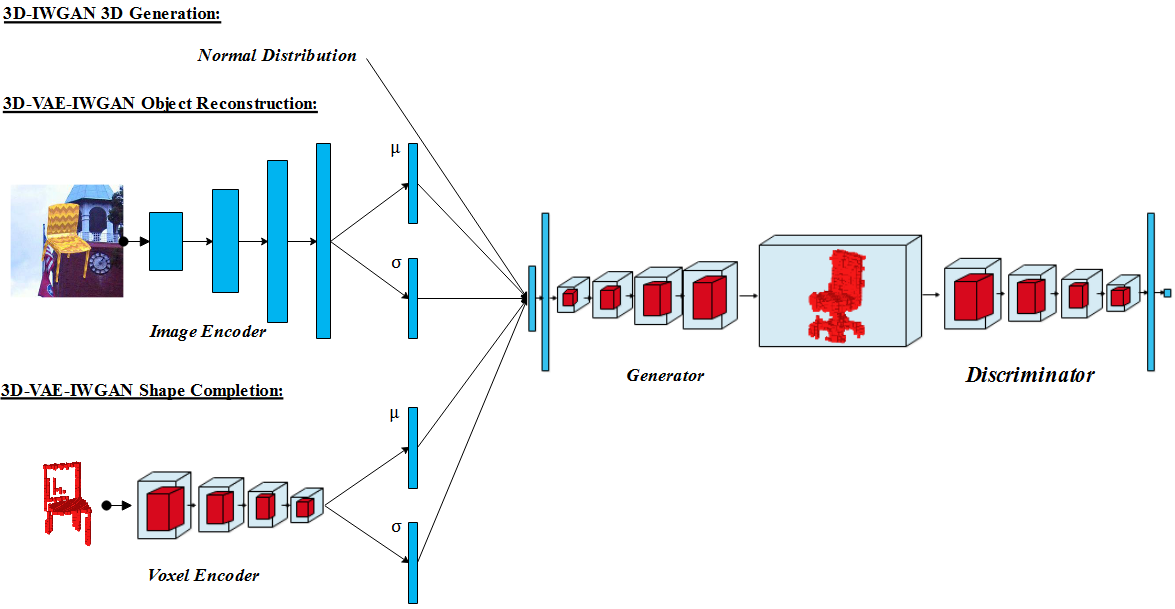}
\caption{A diagram outlining a forward pass though our three 3D generative systems.}
\label{Diagram}
\end{figure} 

\subsection{3D Generation} 
The GAN algorithm is a generative system consisting of two networks, the generator and the discriminator, which learn together to mimic a target distribution \citep{GAN_Original}. The generator converts normally distributed latent vectors into samples which the discriminator then classifies as either generated or real. They play the following minimax game with value function $V(G,D)$: 
\begin{equation}
\min_G \max_D V(D,G) = \mathbb{E}_{x \sim P_{data}} [ \log{D(x)} ] +  \mathbb{E}_{z \sim P_{noise}} [ \log{(1-D(G(z)))} ]
\end{equation}
where $x$ is sampled from the target $P_{data}$ and $z$ is a normally distributed vector. Theoretical results have shown \cite{GAN_Original} that GAN training minimizes the Kullback--Leibler (KL) divergence between the data and generated distributions, with convergence to equality in idealized conditions. When training deep networks in this fashion, the use of gradient descent leads to unstable learning. One must carefully balance the learning of the generator and the discriminator, or gradients can vanish which prevents improvement. The Wasserstein GAN algorithm (WGAN) \citep{Wasserstein} attempts to fix these flaws by instead minimizing the Wasserstein distance between the distributions. This metric is defined for $\prod(P_{r},P_{g})$, the set of all joint distributions whose marginal distributions are $P_r$ and $P_g$, as  
\begin{equation}
W(P_r,P_g) = \inf_{\psi \in  \prod(P_{r},P_{g})} \mathop{\mathbb{E}}_{(x,y) \sim \psi } [|| x-y ||].
\end{equation}
The WGAN scheme has been shown to produced far more stable convergence, and allow one to track convergence through the discriminator's loss \citep{Wasserstein}. The WGAN system's main departure from the original system is clipping the discriminators weights to lie within a compact space, which forces the the discriminator to lie within the set of 1-Lipschitz functions. This constraint is a key in ensuring constructed Wasserstein distance is always continuous and almost always differentiable.

The new Improved Wasserstein GAN training system (IWGAN) \citep{IWGAN} provides a more desirable method for enforcing the Lipschitz constraint. As opposed to weight clipping, it penalizes deviation of the discriminator's gradients from unity, as the gradients of a differentiable function are at most 1 if and only if it is a 1-Lipschitz function. This results in the following loss function for the discriminator: 
\begin{equation}
\mathbb{E}_{\hat{x} \sim P_g} [D({\hat{x}})] - \mathbb{E}_{x \sim P_r} [D({x})] + \lambda  \mathbb{E}_{\hat{x} \sim P_x} [(||\nabla_{\hat x}D(\hat x)||_2 -1)^2]
\end{equation}
where $P_g$ and $P_r$ are the generator and target distribution, and $P_x$ is the distribution sampling uniformly on a straight line between $P_g$ and $P_x$. Gulrajani et al. \citep{IWGAN} shows that weight clipping pushes the networks towards overly simple functions and can often lead to vanishing or exploding gradients. When penalizing the gradient, however, these problems no longer occur, leading to more complex network functions and more stable convergence. We build upon the ideas leveraged in the 3D-GAN system \citep{3DGAN}, applying our own new architectures and substituting the standard GAN training model for the new IWGAN system. 

In our new 3D-IWGAN system, shortcomings of the original system, including a poor ability to generalize to more varied object distributions and unstable convergence are ameliorated. The gradient penalty specified above is added to the discriminator's loss, with $\lambda = 10$. Objects at random points on straight lines between our generated objects and real objects are used for its calculation, as suggested in \citep{IWGAN}. As opposed to the original GAN scheme \citep{GAN_Original}, the generator now learns only every 5 batches, while the discriminator learns on every batch, which leads to more stable convergence. Batch normalization layers are not present in our discriminator as they cause the discriminator's output for each sample to be dependent upon the values of all samples in the observed batch, and so corrupts our gradient penalty. In addition, no activation function is applied to the discriminator's final output, and Adam \cite{ADAM} optimization is used to train both the generator and the discriminator, both with learning rate $10^{-4}$. These changes, which in combination with the network architectures below we refer to as 3D-IWGAN, lead to more stable convergence and allow for quantitative tracking of our models' convergence. 

\subsection{3D Object Reconstruction from Single Images}

We tackle the problem of reconstructing 3D objects from RGB images by combining our new 3D-IWGAN system with the VAE-GAN system \citep{VAE-GAN}, in a novel training scheme. A VAE \citep{VAE} is generative, encoder-decoder system for learning complex distributions. The encoder observes samples from the target distribution and produces a vector of means and variances parameterizing a set of Gaussians, which are sampled to produced a latent vector. This vector is passed to the decoder which attempts to reproduce the original sample. This encoding system is performed to facilitate generation by sampling the latent vector from a normal distribution. Both networks are penalized on the sample's reconstruction error, while the encoder's loss also includes a regularizing KL divergence term encouraging the Gaussians to resemble standard normal distributions. The VAE-GAN system \citep{VAE-GAN} is an algorithm to combine the efficient encoding of VAE's with the generative power of the GAN system to produce samples that can be conditioned on prior data \citep{VAE-GAN}. It works by leveraging a single network for both the generator and decoder, and combining their loss functions.

As is typical for VAE-GAN systems \citep{VAE-GAN}, our VAE's decoder network is simultaneously used for the GAN's generator network. The VAE's encoder converts an image into a 400 dimensional vector of means of variances, which are sampled using Gaussians to produce our latent vector. The latent vector is then passed through the decoder/generator network to generate a reconstructed object, which is then given to the discriminator to pass judgment on its validity. The loss functions for the encoder network and the decoder/generator network are then:

\begin{minipage}{.5\linewidth}

\begin{equation}
 ||\hat x - x||_2 +\KL[N(\mu,\Sigma)||N(0,I)] 
\end{equation}
\end{minipage}%
\begin{minipage}{.4\linewidth}
\begin{equation}
 D({\hat{x}}) + \delta ||\hat x - x||_2
\end{equation}
\end{minipage}

respectively, where x is the target sample, $\hat x$ is the generated sample (generated from an encoded image in the first equation, and a random latent vector in the second) , $\mu$ and $\Sigma$ are the means and variances produced by the encoder, and $\delta = 100$. The discriminator's loss remains the same. 

The same generator and discriminator networks as used for 3D generation are implemented into this system and the encoder network is a simple 5 layer convolution neural network. During training, the discriminator and encoder networks are trained at every batch while the generator only learns every 5 batches. This last point is key to the integration of the systems, as if the encoder is not trained alongside the discriminator at every iteration the system will not converge. This makes sense as both networks should learn similar features about the objects being created at approximately the same rate. We refer to this combined system as 3D-VAE-IWGAN, and a summary diagram of this system can be viewed in Figure \ref{Diagram}. This application is chosen to demonstrate our model's improved generative power, and to introduce our novel system for integrating the Wasserstein Algorithms with the VAE-GAN system. With our new system we achieve state of the art reconstruction results on the IKEA Dataset \citep{IKEA}.

\subsection{Reconstruction of 3D Objects from Single Perspective Depth Scans}
We accurately reconstruct an object's complete 3D shape and volume when presented with only a depth map of the object from a single persecutive, as one might receive from a single snapshot of a Kinect scan. We tackle this problem to show the generative power of our 3D-VAE-IWGAN system, to highlight that our models are applicable to realistic robotic problems, and to demonstrate that our system is easily applicable to tasks involving reproducing 3D shapes from multiple input types.

\subsection{Datasets}
For 3D generation, the ModelNet10 Dataset \citep{ModelNet} was used, consisting of roughly 57000, 32x32x32 resolution objects, from 10 object classes in 12 evenly spaced orientations. The 10 classes are bathtubs, beds, chairs, desks, dressers, monitors, nightstands, sofas, tables and toilets. The goal of our 3D generation system is to be trained on the entire dataset and successfully produce varied objects from all of these classes and orientations simultaneously. 

An entirely synthetic dataset, that we constructed, was exclusively used to train our models for 3D object reconstruction from single images. This dataset consisted of the 6 object classes from the ShapeNet Dataset \citep{ShapeNet}, rendered in front of backgrounds images from the SUN dataset \citep{SUN} at random poses, lighting conditions, and distances,  and overlayed with random textures from the Describable Textures Dataset \citep{Textures}. Accompanying each image was its ground truth object from the ShapeNet Dataset, at 20x20x20 voxel resolution. The IKEA Dataset \cite{IKEA} was used to test our models, which consists of a set of 759 images picturing a large array of objects, and the corresponding object models. These objects fall into 6 categories: beds, bookcases, chairs, desks, sofas, and tables, and are evaluated at resolution 20x20x20. The dataset presents a strong evaluation tool for heavily occluded images in realistic scenes, using only the constraint that the object is centered within the image.   

We constructed a second synthetic dataset, taking 10 random perspectives for each object in the ModelNet10 dataset in order to train models for the shape completion task. A test set of entirely unseen objects was held back for evaluation, examples of which, can be observed in the first rows of Figure \ref{KinectSynth} and Figure \ref{KinectSynth2}. In order to ensure the models trained on our synthetic data was sufficient for application to real world data, a small set of Kinect depth maps sampled from the Large Dataset of Object Scans \citep{Kinect} was converted into voxel representation and also used for evaluation.

\subsection{Network Architectures} 

New network architectures were designed for the generator and the discriminator in our systems to allow for quicker training times, the integration of new GAN techniques, and to conform to the requirements of the IWGAN algorithm. The input to the generator, a 200 dimensional normally distributed latent vector, is passed though a fully connected layer with 2048 nodes, followed by four 3D deconvolutional layers with stride length 2 and kernel size 4. Each deconvolutional layer is followed by a batch normalization layer and a ReLU activation layer, apart from the final layer which is only followed by a tanh activation layer. The output of the network is a 32x32x32 resolution voxel grid.
\begin{figure}[ht]
\includegraphics[width=\textwidth]{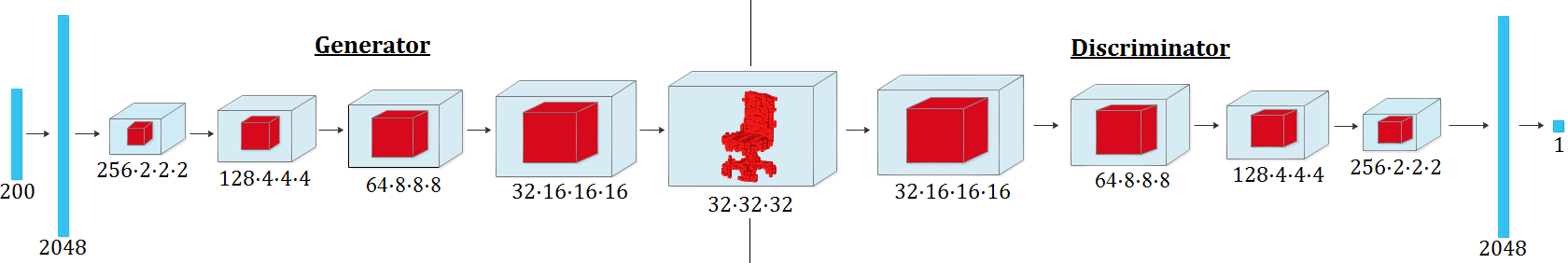}
\caption{The novel network architectures for our generator and discriminator.}
\label{generator}
\end{figure}

The discriminator's network input, a 32x32x32 resolution object, is passed through 4 3D convolutional layers, flattened, and then condensed to a single value through a final fully connected layer. Each convolutional layer posses stride length 2 and kernel size 4, and is followed by a Leaky ReLU activation layer. The network architectures for these two systems can be seen in Figure \ref{generator}. The bases for these two architectures are the DC-GAN networks, due to their success in generating high quality images \citep{DCGAN}. The combination of new training system under the IWGAN algorithm and new architectures described above results in our novel generative model which we refer to as as 3D-IWGAN. Figure \ref{Diagram} summarizes this system.

In order to generate objects at 20x20x20 resolution (needed to test on the IKEA Dataset), slightly downscaled versions of the generator and discriminator network were required, though these changes are very natural. The 3D-VAE-IWGAN image encoder mimics the simple 5 layer convolutional neural network encoder architecture described in \citep{3DGAN}. For the 3D-VAE-IWGAN voxel encoder, an architecture identical to that of the discriminator network described above for 3D generation is used, except that the final layer outputs a 400 dimensional vector of means and variances. Here one can more easily understand why it is necessary for the encoder and the discriminator to learn at the same pace. They receive very similar inputs from which we require that they learn practically the same features and at the same rate, so as to steadily push the generator in the right direction.



	

\section{Experiments}
\label{sec:result}
\subsection{3D Generation}

\begin{figure}[bt]
\includegraphics[width=\textwidth]{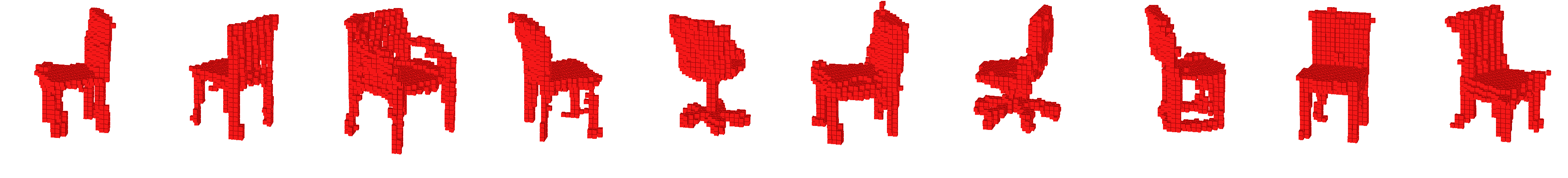}
\caption{Objects generated by the 3D-IWGAN system trained on the ModelNet10 chair class in 12 orientations.}
\label{IWchairs}
\end{figure}

\begin{figure}[ht]
\includegraphics[width=.65\textwidth]{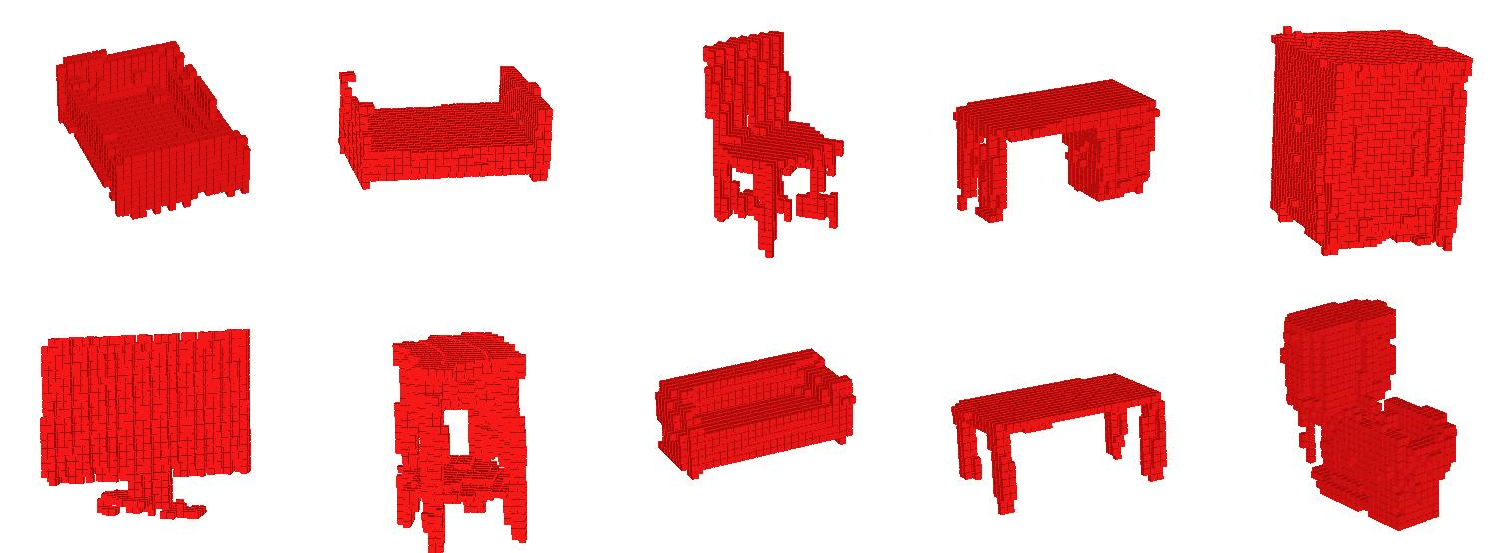}
\centering
\caption{Objects generated by the 3D-IWGAN system, trained on the full ModelNet10 dataset. Each shape above is derived from a different class in the order: bathtub, bed, chair, desk, dresser, monitor, nightstand, sofa, table, and toilet. Their orientations have been altered for optimal viewing.}
\label{IWall}
\end{figure}

\begin{figure}[ht]
\includegraphics[width=.9\textwidth]{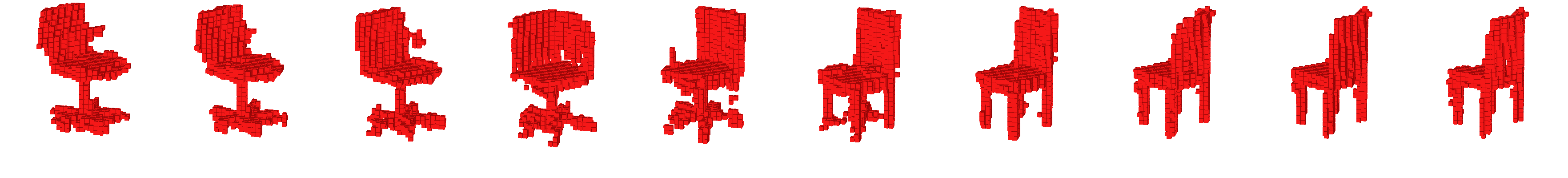} 
\centering
\includegraphics[width=.9\textwidth]{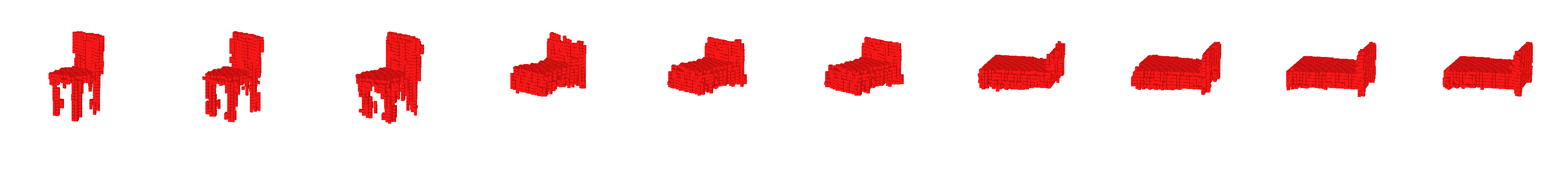}
\centering
\caption{First row: The transition from one chair's latent representation to another, generated by the 3D-IWGAN system when trained on the ModelNet10 chair class. Second Row: The transition from one object's latent representation to another, generated by 3D-IWGAN system when trained on the full ModelNet10 dataset.}
\label{IWinterp}
\label{IWallinterp}
\end{figure}

\begin{figure}[ht]
\includegraphics[width=.5\textwidth]{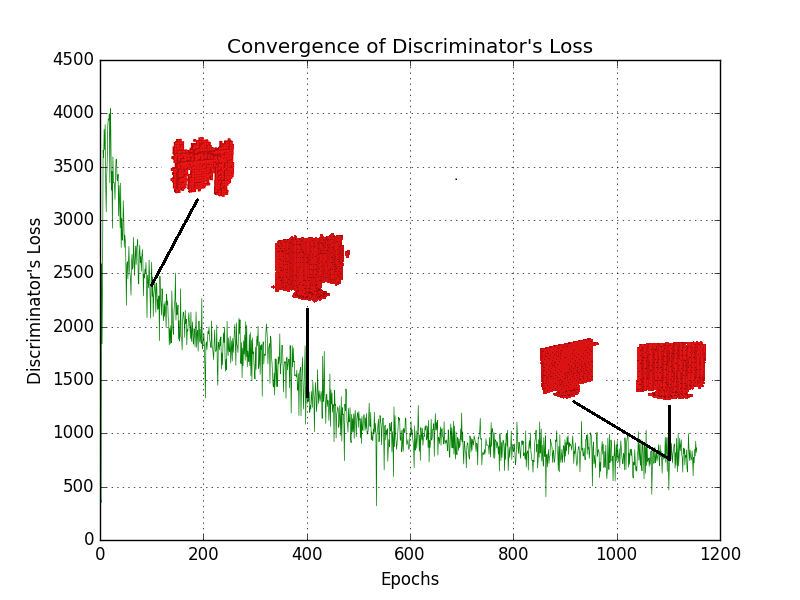}
\centering
\caption{A graph depicting the discriminator's loss at each epoch while training the 3D-IWGAN system on all objects and orientations of the ModelNet10 dataset, along with example monitor objects generated at various points in training.}
\label{IWgraph}
\end{figure}

We first examined if our 3D-IWGAN system was a able to generate objects from a distribution consisting of only chairs, but set in 12 different orientations from the ModelNet10 dataset. As shown in Figure \ref{IWchairs}, our method produces objects of high quality from a variety of viewpoints. Figure \ref{IWinterp} demonstrates that our methods is also able to interpolate between two chairs. As the latent variables shift from the first object to the last, there is not a sudden change, but a slow transition in which each segment can easily be identified as a chair. We next experimented with generating from the entire ModelNet10 dataset, with 10 object classes in 12 orientations simultaneously. This was also a markedly successful endeavor, as the the objects our trained models produced are successfully varied and of high quality. This can be verified in Figure \ref{IWall}. It was possible to track the quality of the objects using the discriminator's loss, which can be viewed in Figure \ref{IWgraph}. This allowed the models convergence to be easily tracked and provided a clear signal for when to halt the learning. For the single object class test, the model trained for 4000 epochs, and for the multi-class test the model was trained for 1100 epochs. 

The models trained here converged easily, with no tuning, and far more stably than with the original GAN framework. An example of this stable convergence can be viewed in Figure \ref{IWgraph}. We can see that the 3D-IWGAN system is a powerful generative model, consistently producing high quality objects, from distinctly difficult distributions.  

To gain some understanding of how our 3D-IWGAN models deal with the transitions from one distinct object class and orientation to another, observe the bottom row of Figure \ref{IWallinterp}. The transitions between the objects are all recognizable and clean. This is a remarkable change from training the original GAN system, which is well-known to induce large unusable areas between properly generated objects which produce unrealistic shapes \citep{Modelling}. Our model's more attractive transitions between objects demonstrate that it has learned a better organization of observed objects' positions along the latent space, so as to produce as few of these unrealistic areas as possible. Our improved training scheme makes the unavoidable unrealistic interpolations represent as small a portion of the latent space as possible. 

\subsection{3D Object Reconstruction from Single Images}
3D-IWGAN was applied to the task of recovering the 3D shape of objects found in images. This was a task also tackled by the 3DGAN system \citep{3DGAN}, and so provides fair and quantitative basis for comparison of the two methods.  The results of our trained 3D-VAE-IWGAN system evaluated on the IKEA dataset can be observed in Table \ref{IKEAtable}. Our system consistently outperforms the original 3DGAN system and several other previous approaches, with a mean average precision of 61.7 across all classes. Our jointly trained model also demonstrates improvement over those from all previous work, on average, and falls only slightly short of our new separately trained models. We note that a joint model faces a harder problem, intrinsically here, as the independent models are provided access to known category labels, while joint models are competitive in the \emph{fully unsupervised} regime. These results demonstrate the large generative improvements which result from our new 3D-IWGAN system and the success of our new 3D-VAE-IWGAN system. 

\begin{table}[ht]
\centering
\caption{Average Precision Scores on the IKEA Dataset}
\label{IKEAtable}
\scalebox{0.8}{
\renewcommand{\arraystretch}{1.2}
\begin{tabular}{|l|llllll|l|}
\hline
Method                                        & Bed           & Bookcase & Chair         & Desk          & Sofa          & Table & Mean \\ \hline
AlexNet-fcc8 \citep{old}                       & 29.5          & 17.3     & 20.4          & 19.7          & 38.8          & 16.0  & 23.6 \\ \hline
AlexNet-conv4 \citep{old}                     & 38.2          & 26.6     & 31.4          & 26.6          & 69.3          & 19.1  & 35.2 \\ \hline
T-L Network \citep{old}                        & 56.3          & 30.2     & 32.9          & 25.8          & 71.7          & 23.3  & 40.0 \\ \hline
3D-VAE-GAN \citep{3DGAN} jointly trained    & 49.1          & 31.9     & 42.6          & 34.8          & 79.8          & 33.1  & 41.2 \\ \hline
3D-VAE-GAN \citep{3DGAN} separately trained & 63.2          & 46.3     & 47.2          & 40.7          & 78.8          & 42.3  & 53.1 \\ \hline
3D-VAE-IWGAN (ours)  jointly trained     &  65.7             & 44.2         & 49.3               &  \textbf{50.6}              & 68.0              & 52.2      & 55.0     \\ \hline
3D-VAE-IWGAN (ours) separately trained   & \textbf{77.7} &    \textbf{51.8}     & \textbf{56.2} &49.8 & \textbf{82.0} & \textbf{52.6}       & \textbf{61.7}      \\ \hline
\end{tabular}}
\end{table}

\subsection{Reconstruction of 3D Objects from Single Perspective Depth Scans}
\begin{figure}[ht]
\centering
\includegraphics[width=.6\textwidth]{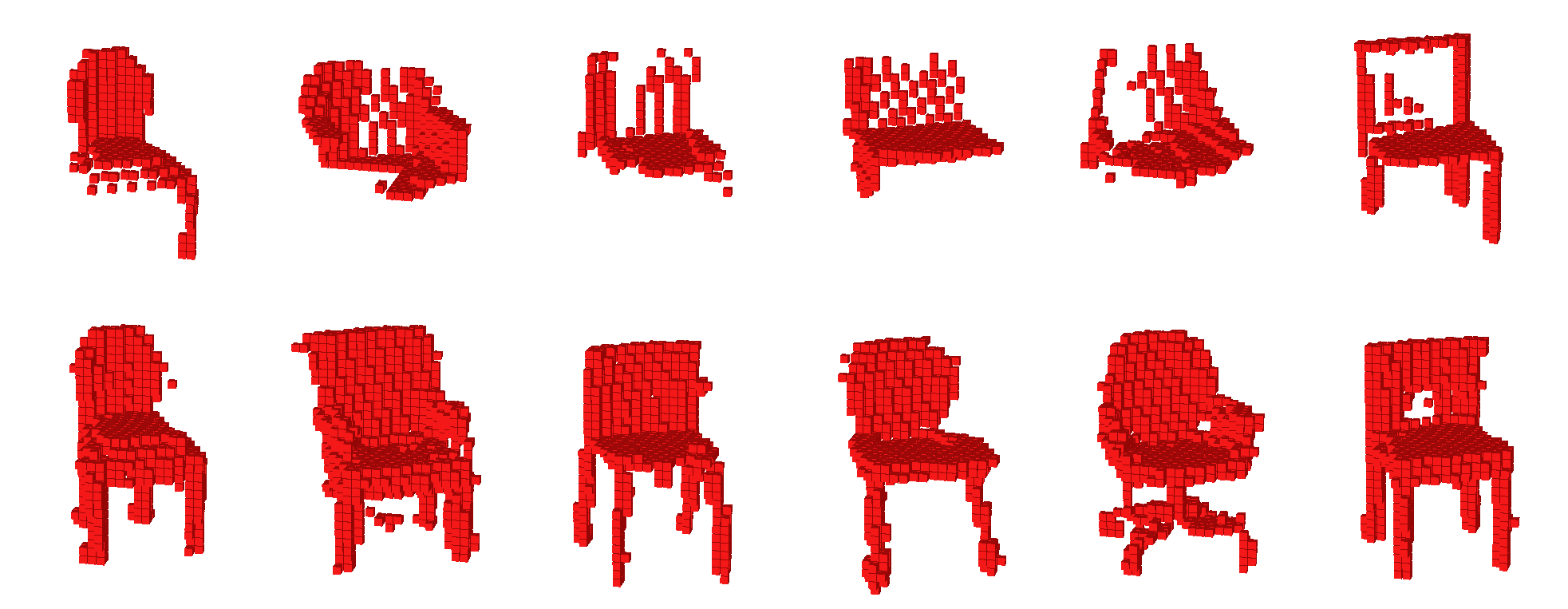}
\caption{First row: Example synthetic single perspective kinetic scans created by the authors, produced from the chair class of the ModelNet10 dataset.  
Second row: The corresponding 3D object reconstruction result of our 3D-IWGAN system.}
\label{KinectSynth}
\end{figure}

\begin{figure}[ht]
\centering
\includegraphics[width=.6\textwidth]{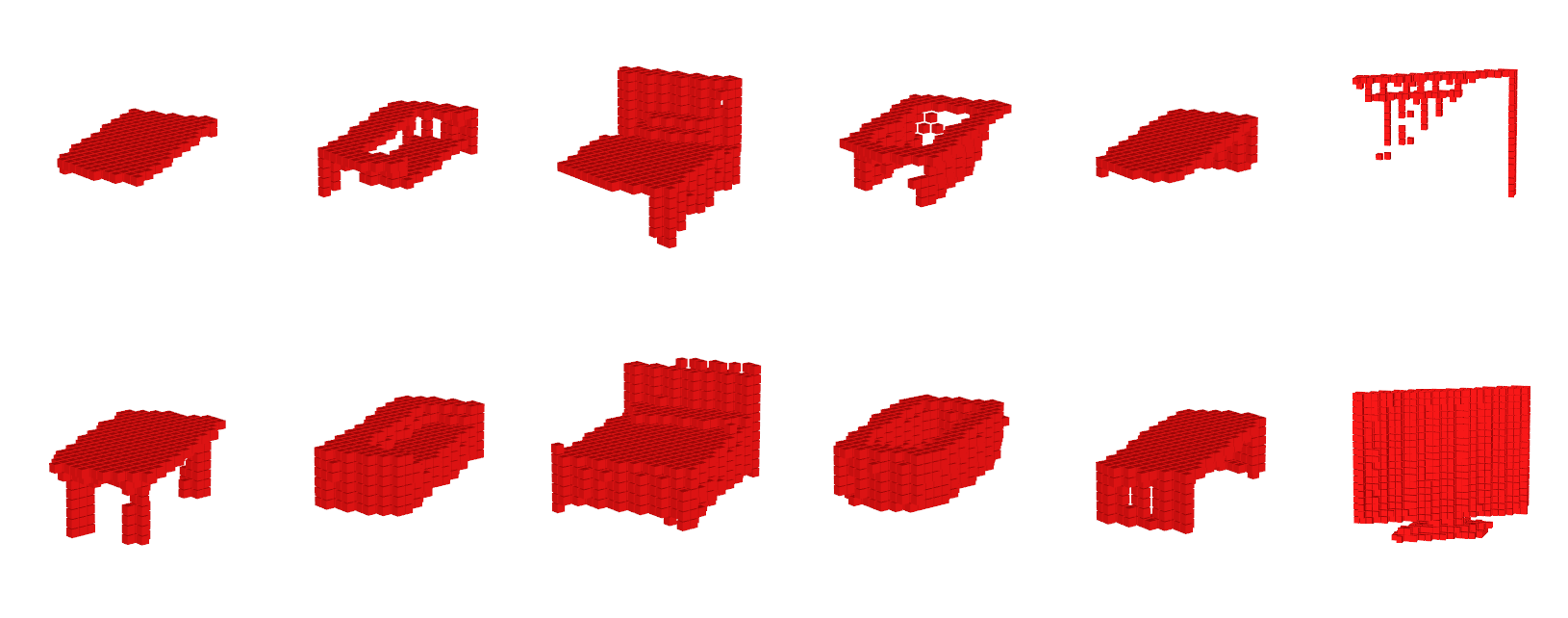}
\caption{First row: Example synthetic single perspective kinetic scans created by the authors, produced from the ModelNet10 dataset.  
Second row: The corresponding 3D object reconstruction result of our 3D-IWGAN system.}
\label{KinectSynth2}
\end{figure}

A voxel encoded variation of our 3D-VAE-IWGAN system was applied to the task of recovering the complete 3D shape of objects from the output of a single perspective scan from a depth sensor. Two models were produced; the first trained on only chair objects and the second on all the objects in the ModelNet10 dataset. It was apparent that both tests were quite successful and examples of recovered objects from the first and second tests set can be viewed in the second rows of Figure \ref{KinectSynth}, and Figure \ref{KinectSynth2}. 

\begin{figure}[ht]

\hbox{\hspace{8.6em} \includegraphics[width=.6\textwidth]{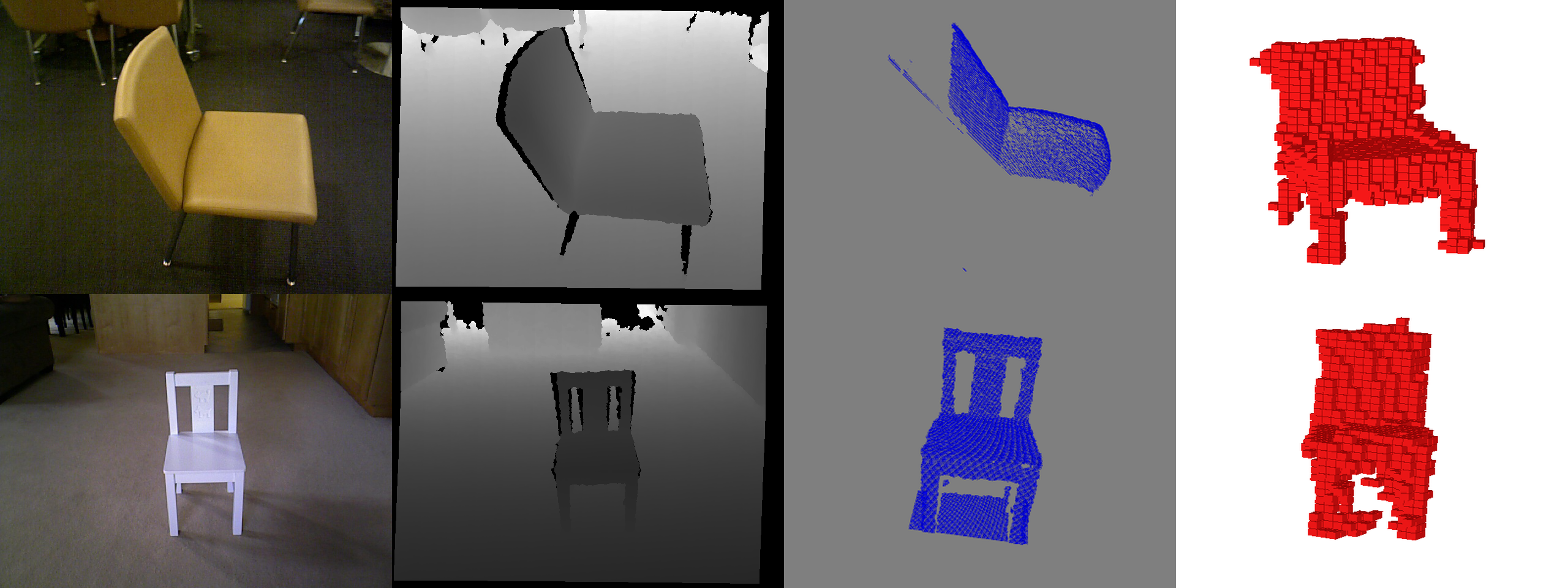}}
\caption{The pipeline converting Kinect RGB-D images to voxel representations and the final object reconstructions produced using the 3D-VAE-IWGAN.}
\label{KinectReal}
\end{figure}

To demonstrate that the models produced here are effective on actual Kinect scans, a set of real scans was evaluated as well. A small sample of Kinect depths maps, found in the \emph{A Large Dataset of Object Scans} dataset \citep{Kinect} was converted into voxel representations, and passed though the trained GAN model for chair reconstruction. Two examples of this pipeline and the objects reconstructed through it can be viewed in Figure \ref{KinectReal}. This picture shows that our generative model accurately reconstructs the original chairs. The success in this task both on the synthetic models and true Kinect scan illustrates that the generative power of our 3D-VAE-IWGAN system makes it is easily applicable to tasks involving reproducing 3D shapes, from a variety of input types. 

\section{Conclusion}
\label{sec:conclusion}
In this paper we outline a new system, 3D-IWGAN, which is successful in 3D generation from complex distributions involving multiple distinct classes; spaces which we highlight other 3D generative systems involving GANs cannot generalize to. We show the models produced by this system learn distributions involving detailed, multiple object classes in multiple orientations, and highlight that these models are easy to train, require no special tuning, and have a quantitative convergence. In addition, we describe how to combine this new system with Variational Auto-encoders to produce a 3D generative system, 3D-VAE-IWGAN, the output of which can be conditioned on some known data. We then demonstrate this system's generative power by recovering 3D objects from images, to achieve state of the art performance on the IKEA dataset. Finally, we again demonstrate our system's generative power and also its ability to be conditioned on complex data by successfully applying it to recovering objects' 3D volume and shape from single perspective depth maps.


\nocite{DBLP:journals/corr/SharmaGF16}
\clearpage


\bibliography{gan}

\end{document}